\documentclass{article}
\usepackage{spconf,amsmath,epsfig}

\usepackage{amsmath,amssymb,amsfonts}
\usepackage{algorithmic}
\usepackage{diagbox}
\usepackage{graphicx}
\usepackage{textcomp}
\usepackage{xcolor}
\usepackage{array}
\usepackage{algorithm}
\usepackage{algorithmic}
\usepackage{bm}
\usepackage{graphicx} %use graph format
\usepackage{epstopdf}
\begin{document}\sloppy

% Example definitions.
% --------------------
\def\x{{\mathbf x}}
\def\L{{\cal L}}

% Title.
% ------
\title{Many could be better than all: A novel instance-oriented algorithm for Multi-modal Multi-label problem}
%
% Single address.
% ---------------
\name{Yi Zhang, ~Cheng Zeng, ~Hao Cheng,~Chongjun Wang, ~Lei Zhang$^{*}$\thanks{$^{*}$Corresponding author}}
\address{National Key Laboratory for Novel Software Technology at Nanjing University \\
%Department of Computer Science and Technology \\
Nanjing University, Nanjing 210023, China \\
\{mg1733091, njuzengc, chengh\}@smail.nju.edu.cn, \{chjwang, zhangl\}@nju.edu.cn}

\maketitle

\begin{abstract}
With the emergence of diverse data collection techniques, objects in real applications can be represented as multi-modal features. What's more, objects may have multiple semantic meanings. Multi-modal and Multi-label \cite{ye2016college} (MMML) problem becomes a universal phenomenon. The quality of data collected from different channels are inconsistent and some of them may not benefit for prediction. In real life, not all the modalities are needed for prediction. As a result, we propose a novel instance-oriented Multi-modal Classifier Chains (MCC) algorithm for MMML problem, which can make convince prediction with partial modalities. MCC extracts different modalities for different instances in the testing phase. Extensive experiments are performed on one real-world herbs dataset and two public datasets to validate our proposed algorithm, which reveals that it may be better to extract many instead of all of the modalities at hand. 
\end{abstract}
\section{Introduction}
\label{sec:intro}

In many natural scenarios, objects might be complicated with multi-modal features and have multiple semantic meanings simultaneously. 

For one thing, data is collected from diverse channels and exhibits heterogeneous properties: each of these domains present different views of the same object, where each modality can have its own individual representation space and semantic meanings. Such forms of data are known as multi-modal data. In a multi-modal setting, different modalities are with various extraction cost. Previous researches, i.e., dimensionality reduction methods, generally assume that all the multi-modal features of test instances have been already extracted without considering the extraction cost. While in practical applications, there is no aforehand multi-modal features prepared, modality extraction need to be performed in the testing phase at first. While for the complex multi-modal data collection nowadays, the heavy computation burden of feature extraction for different modalities has become the dominant factor that hurts the efficiency.

For another, real-world objects might have multiple semantic meanings. To account for the multiple semantic meanings that one real-world object might have, one direct solution is to assign a set of proper labels to the object to explicitly express its semantics. In multi-label learning, each object is associated with a set of labels instead of a single label. Previous researches, i.e., classifier chains algorithm is a high-order approach considering the relationship among labels, but it is affected by the ordering specified by predicted labels. 

To address all the above challenges, this paper introduces a novel algorithm called Multi-modal Classifier Chains (MCC) inspired by Long Short-Term Memory (LSTM) \cite{lstm1}\cite{lstm2}. Information of previous selected modalities can be considered as storing in memory cell. The deep-learning framework simultaneously generates next modality of features and conducts the classification according to the input raw signals in a data-driven way, which could avoid some biases from feature engineering and reduce the mismatch between feature extraction and classifier. 
The main contributions are:
\begin{itemize} 
\item We propose a novel MCC algorithm considering not only interrelation among different modalities, but also relationship among different labels.
\item MCC algorithm utilizes multi-modal information under budget, which shows that MCC can make a convince prediction with less average modality extraction cost.
\end{itemize}
The remainder of this paper is organized as follows. Section 2 introduces related work. Section 3 presents the proposed MCC model. In section 4, empirical evaluations are given to show the superiority of MCC. Finally, section 5 presents conclusion and future work.

\section{Related Work}
In this section, we briefly present state-of-the-art methods in multi-modal and multi-label \cite{multilabel} fields. As for modality extraction in multi-modal learning, it is closely related to feature extraction \cite{featureselection}. Therefore, we briefly review some related work on these two aspects in this section. 

Multi-label learning is a fundamental problem in machine leaning with a wide range of applications. In multi-label learning, each instance is associated with multiple interdependent labels. Binary Relevance (BR) \cite{br} algorithm is the most simple and efficient solution of multi-label algorithms. However, the effectiveness of the resulting approaches might be suboptimal due to the ignorance of label correlations. To tackle this problem, Classifier Chains (CC) \cite{cc} was proposed as a high-order approach to consider correlations between labels. It is obviously that the performance of CC is seriously affected by the training order of labels. To account for the effect of ordering, Ensembles of Classifiers Chains (ECC) \cite{cc} is an ensemble framework of CC, which can be built with $n$ random permutation instead of inducing one classifier chain. Entropy Chain Classifier (ETCC) \cite{etcc} extends CC by calculating the contribution between two labels using information entropy theory while Latent Dirichlet Allocation Multi-Label (LDAML) \cite{ldaml} exploiting global correlations among labels. LDAML mainly solve the problem of large portion of single label instance in some special multi-label datasets. Due to high dimensionality of data , dimensionality reduction \cite{dimensionreduction} or feature extraction should be taken into consideration. 

Originally, feature selection and dimensionality reduction are generally used for reducing the cost of feature extraction. \cite{songandlu} proposed regularized multilinear regression and selection for automatically selecting a set of features while optimizing prediction for high-dimensional data. Feature selection algorithms do not alter the original representation of the variables, but merely select part of them. Most of existing multi-label feature selection algorithms either boil down to solving multiple single-label feature selection problems or directly make use of imperfect labels. Therefore, they may not be able to find discriminative features that are shared by multiple labels. To reduce the negative effects of imperfect label information in finding label correlations, \cite{mifs} decomposes the multi-label information into a low-dimensional space and then employs the reduced space to steer the feature selection process. Furthermore, we always extract multiple features rather than single feature for classification, several adaptive decision methods for multi-modal feature extraction are proposed \cite{lp} \cite{dag}. To further reduce the number of features for testing, \cite{dmp} proposes a novel Discriminative Modal Pursuit (DMP) approach. 

In this paper, taking both multi-label learning and feature extraction into consideration, we propose MCC model with an end-to-end approach \cite{end2end} for MMML problem, which is inspired by adaptive decision methods. 
%Specially, with the given raw training data with multi-modal features, we learn an instance specific discriminative modalities of features extraction method. 
Different from previous feature selection or dimensionality reduction methods, MCC extracts different modalities for different instances and different labels. Consequently, when presented with an unseen instance, we would extract the most informative and cost-effective modalities for it. Empirical study shows the efficiency and effectiveness of MCC, which can achieve better classification performance with less average modalities.

\section{Methodology}
\begin{figure}[!htbp]
\centerline{\includegraphics[height=6cm, width=9cm, angle=0]{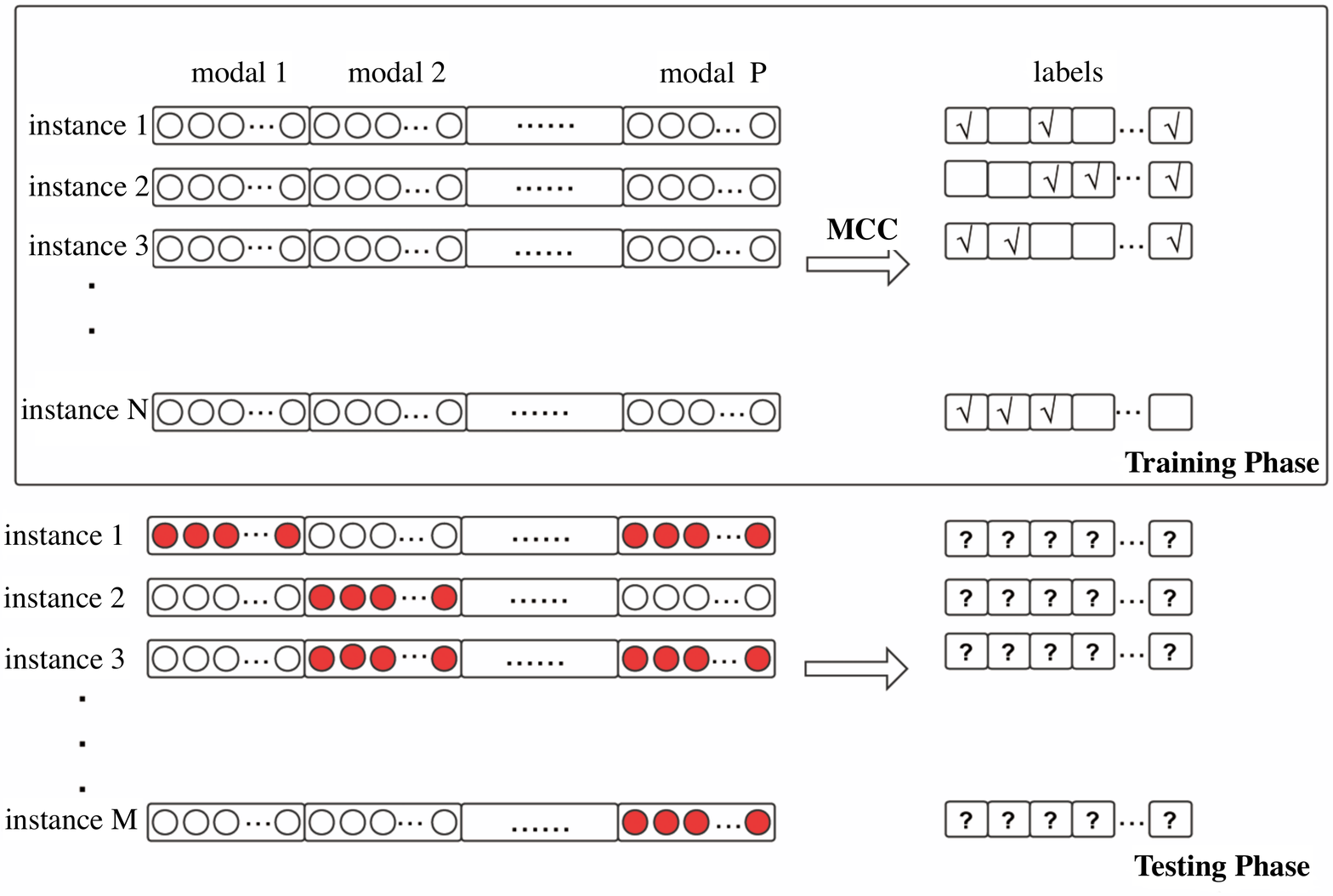}}
\caption{Diagrammatic illustration of MCC model. In the testing phase, circles shadowed with red represent the features used for categorization prediction.}
% MCC adopts different modalities for different instances.
\label{herb}
\end{figure}

This section first summarizes some formal symbols and definitions used throughout this paper, and then introduces the formulation of the proposed MCC model. An overview of our MCC algorithm is shown in Fig.\ref{herb}.
\subsection{Notation}
In the following, bold character denotes vector (e.g., $\bm X$).

The task of this paper is to learn a function $\mathit{h}$: $\mathcal{X}\rightarrow 2^{\mathcal{Y}}$ from a training dataset with $N$ data samples $\mathcal{D}=\{(\bm X_{i}, \bm Y_{i})\}_{i=1}^{N}$. 
The $i$-th instance $(\bm X_{i}, \bm Y_{i})$ contains a feature vector $\bm X_{i} \in \mathcal{X}$ and a label vector $\bm Y_{i} \in \mathcal{Y}$.

$\bm{X}_{i} = [\bm{X}_{i}^{1}, \bm{X}_{i}^{2}, \dots, \bm{X}_{i}^{P}] \in \mathbb{R}^{d_{1}+d_{2}+\dots+d_{P}}$ is a combination of all modalities and $d_{m}$ is the dimensionality of features in $m$-th modality. $\bm{Y}_{i} = [y_{i}^{1}, y_{i}^{2}, \dots, y_{i}^{L}] \in \{-1, 1\}^{L}$ denotes the label vector of $\bm{X}_{i}$.
$P$ is the number of modalities and $L$ is the number of labels.

Moreover, we define $\bm{c} = \{c_{1}, c_{2}, \dots, c_{P}\}$ to represent the extraction cost of $P$ modalities. 
Modality extraction sequence of $\bm{X}_{i}$ is denoted as $\bm S_{i}=\{S_{i}^{1}, S_{i}^{2}, \dots, S_{i}^{m}\}, m \in \{1, 2, \dots, P\}, m\leq P$, where $S_{i}^{m}\in \{1, 2, \dots, P\}$ represents $m$-th modality of features to extract of $\bm X_{i}$ and satisfies the following condition:  $\forall m, n (m \neq n) \in \{1, 2, \dots, P\}, S_{i}^{m} \neq S_{i}^{n}$. It is noteworthy that different instances not only correspond to different extraction sequences but also may have different length of modalities of features extraction sequence.
Furthermore, we define some notations used for testing phase.
Suppose there is a testing dataset with $M$ data samples $\mathcal{T} = \{(\bm X_{i}, \bm Y_{i})\}_{i=1}^{M}.$
We denote predicted labels of $\mathcal{T}$ as $\bm Z =\{ \bm Z_{i}\}_{i=1}^{M}$, in which $\bm Z_{i}=(z_{i}^{1}, z_{i}^{2}, \dots, z_{i}^{L})$ represents all predicted labels of $\bm X_{i}$ in $\mathcal{T}$ and $\bm Z^{j} = (z_{1}^{j}, z_{2}^{j}, \dots, z_{M}^{j})^{T}$ represents $j$-th predicted labels of all testing dataset.

\subsection{MCC algorithm}
On one hand, MMML is related to multi-label learning and here we extend Classifier Chains to deal with it. On the other hand, each binary classification problem in Classifier Chains can be transferred into multi-modal problem and this procedure aims at making a convince prediction with less average modality extraction cost.

% \begin{figure*}[htbp]
% \centerline{\includegraphics[height=4cm, width=16cm, angle=0]{dsfe.eps}}
% \caption{Comparison of feature numbers used in test phase among Disjoint Set of Features Extraction, Feature Selection and Dimensionality Reduction. In the figure above, circles shadowed with colors indicate features used in the test phase.}
% \label{dsfe}
% \end{figure*}

\subsubsection{Classifier Chains}
Considering correlation among labels, we extend Classifier Chains to deal with Multi-modal and Multi-label problem. Classifier Chains algorithm transforms the multi-label learning problem into a chain of binary classification problems, where subsequent binary classifiers in the chain is built upon the predictions of preceding ones \cite{multilabel}, thus to consider the full relativity of the label hereby. 
The greatest challenge to CC is how to form a recurrence relation chain $\bm \tau$. In this paper, we propose a heuristic Gini index based Classifier Chains algorithm to specify $\bm \tau$. 

First of all, we split the multi-label dataset into several single-label datasets, i.e, for $j$-th label in $\{y^{1}, y^{2}, \dots, y^{L}\}$, we rebuild dataset $\mathcal{D}_{j} = \{(\bm X_{i}, y_{i}^{j})\}$$_{i=1}^{N}$ as $j$-th dataset of single-label. Secondly, we calculate Gini index \cite{gini} of each rebuilt single-label dataset $\mathcal{D}_{j}, (j=1, 2, \dots, L)$. 

\begin{equation}
\label{Gini}
Gini(\mathcal{D}_{j}) = \sum_{k=1}^{|\mathcal{Y}|}\sum_{k\neq k'}p_{k}p_{k'} = 1 - \sum_{k=1}^{|\mathcal{Y}|}p_{k}^{2}
\end{equation}
where $p_{k}$ represents the probability of randomly choosing two samples with same labels, $p_{k'}$ represents the probability of randomly choosing two samples with different labels and $|\mathcal Y|$ represents number of labels in $\mathcal D_{j}$.

And then we get predicted label chain $\bm \tau = \{\tau_{i}\}_{i=1}^{L}$, composed of indexes of sorted $\{Gini(\mathcal D_{i})\}_{i=1}^{L}$ which is sorted in descending order. For $L$ class labels $\{y^{1}, y^{2}, \dots, y^{L}\}$, we are supposed to split the label set one by one according to \boldmath$\tau $ \unboldmath and then train $L$ binary classifiers. 

For the $j$-th label $y^{\tau_j}, (j = 1, 2, \dots, L)$ in the ordered list $\bm \tau$, a corresponding binary training dataset is reconstructed by appending a set of labels preceding $y_{i}^{\tau_{j}}$ to each instance $\bm X_{i}$: 
\begin{equation}
\label{trainingdata}
\mathcal{D}_{\tau_j}=\{([\bm X_{i},\bm{xd}_{i}^{\tau_{j}}], y_{i}^{\tau_{j}})\}_{i=1}^{N}
\end{equation}
where $\bm{xd}_{i}^{\tau_{j}} = (y_{i}^{\tau_{1}}, \dots, y_{i}^{\tau_{j-1}})$ represents the binary assignment of those labels preceding $y_{i}^{\tau_{j}}$ on $\bm X_{i}$ (specifically $\bm {xd}_{i, \tau_{1}} = \emptyset)$ and $[\bm X_{i},\bm{xd}_{i}^{\tau_{j}}]$ represents concatenating vector $\bm X_{i}$ and $\bm{xd}_{i}^{\tau_{j}}$.  
We denote $c_{l}$ as extraction cost of $\bm{xd}_{i}^{\tau_{j}}$. 
If $j > 1$, we combine the new set $\bm{xd}_{i}^{\tau_{j}}$ as a new modality in $\mathcal D_{\tau_{j}}$. Moreover, each instance in $\mathcal D_{\tau_{j}}$ is composed of $P+1$ modalities of features and extraction cost needs to be updated by appending $c_{l}$ to $\bm{c}$. 
We denote the new extraction cost as $\bm{c}'$ and $\bm{c}'=[\bm{c}, c_{l}]$.

Meanwhile, a corresponding binary testing dataset is constructed by appending each instance with its relevance to those labels preceding $y^{\tau_j}$: 
\begin{equation}
\label{testingdata}
\mathcal{T}_{\tau_j}=\{([\bm X_{i},\bm{xt}_{i}^{\tau_{j}}], y_{i}^{\tau_{j}})\}_{i=1}^{M}
\end{equation}
where $\bm {xt}_{i}^{\tau_{j}} = (z_{i}^{\tau_{1}}, \dots, z_{i}^{\tau_{j-1}})$ represents the binary assignment of those labels preceding $z_{i}^{\tau_{j}}$ on $\bm X_{i}$ (specifically $\bm {xt}_{i}^{\tau_1} = \emptyset)$ and $[\bm X_{i},\bm{xt}_{i}^{\tau_{j}}]$ represents concatenating vector $\bm X_{i}$ and $\bm{xt}_{i}^{\tau_{j}}$. 
We denote $c_{l}$ as extraction cost of $\bm{xt}_{i}^{\tau_{j}}$, which is the same as extraction cost of $\bm{xd}_{i}^{\tau_{j}}$.
If $j>1$, each instance in $\mathcal T_{\tau_{j}}$ is composed of $P+1$ modalities of features and one label $y_{i}^{\tau_{j}}$.

After that, we propose an efficient Multi-modal Classifier Chains (MCC) algorithm, which will be introduced in the following paragraph. By passing a combination of training dataset $\mathcal{D}_{\tau_{j}}$ and extraction cost $\bm{c}'$ as parameters of MCC, we get $\bm Z^{\tau_{j}}$. The final predicted labels of $\mathcal{T}$ is the concatenation of $\bm Z^{\tau_{j}}, (j=1, 2, \dots, L)$, i.e., $\bm Z = (\bm Z^{\tau_{1}}, \bm Z^{\tau_{2}}, \dots, \bm Z^{\tau_{L}})$
% we get a binary classifier $\mathit{f}$ to determine whether $y_{\tau(j)}$ in the instance of testing dataset $\mathcal{DT}_{j}$ is relevant label or not.

\subsubsection{Multi-modal Classifier Chains}
In order to induce a binary classifier $\mathit{f_{l}}: \mathcal{X} \times \{-1, 1\}$ with less average modality extraction cost and better performance in MCC, we design Multi-modal Classifier Chains (MCC) algorithm which is inspired by LSTM. 
MCC extracts modalities of features one by one until it's able to make a confident prediction. MCC algorithm extracts different modalities sequence with different length for difference instances, while previous feature extraction method extract all modalities of features and use the same features for all instances.

MCC adopts LSTM network to convert the variable $\bm X'_{i} \in \mathcal{X}$ into a set of hidden representations $\bm H_{i}^{t} = [\bm h_{i}^{1}, \bm h_{i}^{2}, \dots, \bm h_{i}^{t}]$, $\bm h_{i}^{t} \in \mathbb{R}^{h}$. Here, \boldmath $\hat X$\unboldmath$_{i}^{S_{i}^{t}} = [$\boldmath $\hat {X}$\unboldmath$_{i}^{1},\dots, $\boldmath $\hat X$\unboldmath$_{i}^{m}, \dots, $\boldmath$\hat X$\unboldmath$_{i}^{P}]$ is an adaptation of $\bm X_{i}$. In the $t$-th step, the modality to be extracted is denoted as $S_{i}^{t}$. If $m = S_{i}^{t}$, $\bm{\hat X}_{i}^{m}= \bm X_{i}^{S_{i}^{t}}$, $\bm 0$  otherwise.
% For $\bm X_{i}^{j}$ in $\bm X_{i}$, if $j = S_{i}^{t}$ then \boldmath$\hat X$\unboldmath$_{i}^{j} = \bm X_{i}^{S_{i}^{t}}$, else \boldmath$\hat X$\unboldmath$_{i}^{j}= \bm 0$.
 For example, if $S_{i}^{t} = 3$, \boldmath$\hat X$\unboldmath$_{i}^{3} = [\bm 0, \bm 0, \bm X_{i}^{3}, \dots, \bm 0]$.

Similar to peephole LSTM, MCC has three gates as well as two states:
%%Three gates coordinate with each other to protect and control the cell state, all composed of a sigmoid neural net layer and a pointwise multiplication operation. The sigmoid function $\sigma$ outputs numbers between 0 and 1, describing how much of each component should be let through. A value of 0 means `let nothing through' while a value of 1 means `let everything through'. Here we denote $\bm X'_{t}$ as input features at $t$-th step. 
forget gate layer, input gate layer, cell state layer, output gate layer, hidden state layer, listed as follows:

$\bm f_{t} = \sigma([\bm W_{fc}, \bm W_{fh}, \bm W_{fx}][\bm C_{t-1}, \bm h_{t-1}, $\boldmath$\hat X$\unboldmath$_{t}]^{T} + \bm b_{f})$

$\bm i_{t} = \sigma([\bm W_{ic}, \bm W_{ih}, \bm W_{ix}][\bm C_{t-1}, \bm h_{t-1}, $\boldmath$\hat X$\unboldmath$_{t}]^{T}+ \bm b_{i})$

$\bm C_{t} = \bm f_{t}\cdot \bm C_{t-1} + \bm i_{t}\cdot tanh([\bm W_{ch}, \bm W_{cx}][\bm h_{t-1}, $\boldmath$\hat X$\unboldmath$_{t}]^{T} + \bm b_{C})$

$\bm o_{t} = \sigma ([\bm W_{oc}, \bm W_{oh}, \bm W_{ox}][\bm C_{t}, \bm h_{t-1}, $\boldmath$\hat X$\unboldmath$_t]^{T} + \bm b_{o})$

$\bm h_{t} = \bm o_{t}\cdot tanh(\bm C_{t})$

Different from LSTM, MCC adds two full connections to predict current label and next modality to be extracted. For one thing, there is a full connection between hidden layer and label prediction layer, with weight vector \boldmath$\hat W$\unboldmath$_{l}$. For another, there is a full connection between hidden layer and modality prediction layer, with weight vector \boldmath$\hat W$\unboldmath$_{m}$. Moreover, bias vector are denoted as $\bm b_{l}$ and $\bm b_{m}$ respectively.

\begin{itemize}
\item Label prediction layer: This layer predicts label according to a nonlinear softmax function $\mathit fl^{j}(.)$. 
\begin{equation}
\label{fl}
\mathit fl^{j}(\bm H_{i}^{t}) = \sigma(\bm H_{i}^{t}\bm{ \hat W}_{l} + \bm b_{l})
\end{equation}

\item Modality prediction layer: This layer predicts next modality according to a linear function $fm^{j}(.)$ and selects maximum as next modality to be extracted.
\begin{equation}
\label{fm}
\mathit fm^{j}(\bm H_{i}^{t}) = \bm H_{i}^{t} \bm {\hat W}_{m} + \bm b_{m}
\end{equation}

\end{itemize}

We use $\mathcal FL=[fl^{1}, fl^{2}, \dots, fl^{L}]$ and $\mathcal FM=[fm^{1}, fm^{2}, \dots, fm^{L}]$ to denote the label prediction function set and modality prediction function set respectively.

Next, we design loss function composed of loss term and regularization term for producing optimum and faster results. Above all, we design loss of instance \boldmath$\hat X$\unboldmath$_{i}$ with $S_{i}^{t}$ modality as follows.
\begin{equation}
L_{i}^{t} = L_{l}(\mathit fl^{j}(\bm H_{i}^{t}), y_{i}) + L_{m}(\mathit fm^{j}(\bm H_{i}^{t}), \bm {\hat X }_{i}^{t})
\end{equation}
Here we adopt log loss for label prediction loss function $L_{l}$ and hinge loss for modality prediction loss function $L_{m}$, where modality prediction is measured by distances to K Nearest Neighbors \cite{begum2015data}.

Meanwhile, we add Ridge Regression (L2 norm) to the overall loss function.
\begin{equation}
\begin{split}
\Omega_{i}^{t} = ||\bm {\hat W}_{m}||^{2}+||\bm {\hat W}_{l}||^{2}+ ||\bm c \cdot \mathit fm^{j}(\bm H_{i}^{t})||
\end{split}
\end{equation}
where $||.||$ represents L2 norm and $\bm c$ represents extraction cost of each modality.

The loss term is the sum of loss in all instances at $t$-th step. The overall loss function is as follows.
\begin{equation}
\label{loss}
\begin{split}
\bm L_{t} = \sum_{i}^{N}(L_{i}^{t} + \lambda \cdot \Omega_{i}^{t})
\end{split}
\end{equation}
where $\lambda=0.1$ is trade-off between loss and regularization.

In order to optimize the aforementioned loss function $\bm L_{t}$, we adopt a novel pre-dimension learning rate method for gradient descent called AdaDelta \cite{adadelta}. Here, we denote all the parameters in Eq.\ref{loss} as $\bm W = [$\boldmath $\hat W$\unboldmath$_{m}, $\boldmath $\hat W$\unboldmath$_{l}, \lambda]$.

At $t$-th step, we start by computing gradient $\bm g_{t} = \frac{\partial \bm L_{t}}{\partial \bm W_{t}}$ and accumulating decaying average of the squared gradients:
\begin{equation}
\label{eg2}
E[\bm g^{2}]_{t} = \rho E[\bm g^{2}]_{t-1}+(1-\rho)\bm g_{t}^{2}
\end{equation}
where $\rho$ is a decay constant and $\rho = 0.95$.

\begin{algorithm}[ht] 
\caption{The pseudo code of MCC algorithm} 
\label{MCCAlgorithm} 
\begin{algorithmic}[1] %这个1 表示每一行都显示数字
\REQUIRE ~~\\ %算法的输入参数：Input
$\mathcal{D}=$\{(\boldmath$X$\unboldmath$_{i}, \bm Y_{i})\}_{i=1}^{N}$: Training dataset;\\ 
$\bm{c}=$\{$c_{i}\}_{i=1}^{P}$: Extraction cost of $P$ modalities;\\
% $\mathcal{T}=$\{(\boldmath$X$\unboldmath$_{i}, \bm Y_{i})\}_{i=1}^{M}$: Testing dataset;\\
\ENSURE ~~\\ %算法的输出：Output
% $\bm z = \{z_{i}\}_{i=1}^{M}$: Predicted label of $\mathcal{T}$
$\mathcal{FL}:$ set of label prediction function\\
$\mathcal{FM}:$ set of modality prediction function
\STATE Calculate predicted label chain $\bm \tau = \{\tau_{i}\}_{i=1}^{L}$ with Eq.\ref{Gini}
\FOR {$j$ in $\bm \tau$}
\STATE Construct $\mathcal D_{\tau_{j}}$ with Eq.\ref{trainingdata}
\WHILE{$cnt<N_{iter}, cnt$++}
\STATE Initial $E[\bm g^{2}]_{0} = E[\bm \triangle \bm W^{2}]_{0}=0$ 
\STATE Choose $N_{b}$ samples in $\mathcal D_{{\tau_{j}}}$
\FOR{$i = 1:N_{b}$}
\FOR{$t = 1:P$}
\STATE Select $S_{i}^{t}$ with Eq.\ref{fm} and calculate \boldmath$\hat X$\unboldmath$_{i}^{S_{i}^{t}}$
\STATE $\hat c_{i}^{t} = \hat c_{i}^{t} + c_{S_{i}^{t}}$
\IF{$\hat c_{i}^{t} > C_{th}$ or $a_{i}^{t} > A_{th}$ }
\STATE break
\ENDIF
\STATE Calculate $L_{t}$ with Eq.\ref{loss}
\STATE Compute gradient $\bm g_{t} = \frac{\partial \bm L_{t}}{\partial \bm W_{t}}$
\STATE Accumulate gradient $E[\bm g^{2}]_{t}$ with Eq.\ref{eg2}
\STATE Compute Update $\triangle \bm W_{t}$ with Eq.\ref{deltaw}
\STATE Accumulate Updates $E[\bm \triangle \bm W^{2}]_{t}$ with Eq.\ref{edeltaw}
\STATE Update $\bm W_{t+1} = \bm W_{t} +  \triangle \bm W_{t}$
%\STATE \quad if $c_{sum} > c_{th}$ or $con^{i}_{t} > con_{th}$ then
%\STATE \quad break
\ENDFOR
\ENDFOR
\ENDWHILE
\STATE Update $fl^{j}$ and $fm^{j}$ as in Eq.\ref{fl} and Eq.\ref{fm}
\ENDFOR
\RETURN $\mathcal {FL}, \mathcal {FM}$; %算法的返回值
\end{algorithmic}
\end{algorithm}

The resulting parameter update is then: 
\begin{equation}
\label{deltaw}
\triangle \bm W_{t} = -\frac{\sqrt{E[(\triangle \bm W)^{2}]_{t-1}+\epsilon}}{\sqrt{E[\bm g^{2}]_{t}+\epsilon}}\bm g_{t}
\end{equation}
where $\epsilon$ is a constant and $\epsilon = 1e^{-8}$.

And then, we accumulate update:
\begin{equation}
\label{edeltaw}
E[\bm \triangle \bm W^{2}]_{t} = \rho E[\bm \triangle \bm W^{2}]_{t-1}+(1-\rho)\bm \triangle \bm W_{t}^{2}
\end{equation}

The pseudo-code of MCC is summarized in Algorithm \ref{MCCAlgorithm}. $N_{b}$ denotes batch size of training phase. $N_{iter}$ represents maximum number of iterations. $C_{th}$ represents the threshold of cost. $A_{th}$ represents the threshold of accuracy of the predicted label. $\hat c_{i}^{t}$ denotes the sum of extraction cost and $a_{i}^{t}$ denotes accuracy of current predicted label.

\section{Experiment}
% In order to validate the effectiveness of our proposed MCC algorithm, we compare MCC with both multi-label and multi-modal learning methods on one real-world dataset and two public datasets as benchmarks.
\subsection{Dataset Description}
We manually collect one real-world \textit{Herbs} dataset and adapt two publicly available datasets including \textit{Emotions} \cite{emotions} and \textit{Scene} \cite{br}. As for \textit{Herbs}, there are 5 modalities with explicit modal partitions: channel tropism, symptom, function, dosage and flavor. As for \textit{Emotions} and \textit{Scene}, we divide the features into different modalities according to information entropy gain. The details are summarized in Table \ref{datasets}. 
\begin{table}[ht]
\caption{Datasets description. $N$, $L$ and $P$ denote the number of instances, labels and modalities in each dataset, respectively. $D$ shows the dimensionality of each modality.}\label{datasets}
\begin{center}
%\begin{tabular}{p{1.3cm}<{\centering}| p{2cm}<{\centering}|p{2cm}<{\centering}| p{2cm}<{\centering}}
\begin{tabular}{c c c c c }
\hline
\textbf{Datasets}& $N$ & $L$ & $P$ &$D$ \\
\cline{1-4}
\hline
\it{Herbs} & 11104 & 29 & 5 & [13, 653, 433, 768, 36]\\
\it{Emotions} & 593 & 6 & 3 & [32, 32, 8]\\
\it{Scene} & 2407 & 6 & 6 &[49, 49, 49, 49, 49, 49]\\
\hline
\end{tabular}
\label{datasets}
\end{center}
\end{table}
\subsection{Experimental Settings}
All the experiments are running on a machine with 3.2GHz Inter Core i7 processor and 64GB main memory.

We compare MCC with four multi-label algorithms: BR, CC, ECC, MLKNN\cite{mlknn} and one state-of-the-art multi-modal algorithm: DMP\cite{dmp}. 
For multi-label learner, all modalities of a dataset are concatenated together as a single modal input. For multi-modal method, we treat each label independently.

F-measure is one of the most popular metrics for evaluation of binary classification\cite{fmeasure2}.
%Micro-average based on F-measure is widely used to measure the performance of multi-label classification algorithms.
% and it is defined as the harmonic mean of precision and recall.
To have a fair comparison, we employ three widely adopted standard metrics, i.e., Micro-average, Hamming-Loss, Subset-Accuracy\cite{multilabel}. In addition, we use Cost-average to measure the average modality extraction cost.  
%Cost-average evaluates the average extraction cost used for each instance.
For the sake of convenience in the regularization function computation, extraction cost of each modality is set to $1.0$ in the experiment. Furthermore, we set the cost of new modality (predicted labels) to $0.1$ to demonstrate its superiority compared with DMP. 

\subsection{Experimental Results}\label{AA}
For all these algorithms, we report the best results of the optimal parameters in terms of classification performance. Meanwhile, we perform 10-fold cross validation (CV) and take the average value of the results in the end. 

For one thing, table \ref{experimental result} shows the experimental results of our proposed MCC algorithm as well as other five comparing algorithms. 
%%Taking herbs dataset as an example, comparing MCC with ECC algorithm, Micro-average increases by 3.2$\%$, Macro-average increases by 2.19$\%$, Subset-Accuracy increases 7.3$\%$, Hamming-Loss decreases by 0.43$\%$.
%%Furthermore, MCC reduces cost-average by 2.1 compared with other four algorithms.
\begin{table}[ht]
\caption{Comparison results (mean$\pm$std). $\uparrow/\downarrow$ indicates that the larger/smaller the better of a criterion. The best performance on each dataset is bolded.}\label{experimental result}
\begin{center}
\begin{tabular}{p{1.3cm}<{\centering} p{2cm}<{\centering}p{2cm}<{\centering} p{2cm}<{\centering}}
%\begin{tabular}{|c | c| c| c| }
\hline
\textbf{Algorithm}&\multicolumn{3}{c}{\textbf{Evaluation Metrics}} \\
\cline{1-4} 
\textbf{} & {{Micro-average}$\uparrow$}& {{Hamming-Loss}$\downarrow$}& {{Subset-Accuracy}$\uparrow$} \\
\hline
\textit{Herbs}&\multicolumn{3}{c}{\textbf{}} \\
\hline
BR& 0.621$\pm$0.061& 0.033$\pm$0.004& 0.349$\pm$0.060\\
CC& 0.624$\pm$0.060& 0.033$\pm$0.004& 0.363$\pm$0.072\\
ECC& 0.675$\pm$0.010& 0.035$\pm$0.001& 0.376$\pm$0.018\\
MLKNN& 0.544$\pm$0.047& 0.039$\pm$0.005& 0.281$\pm$0.068\\
DMP& 0.635$\pm$0.073& 0.032$\pm$0.004& 0.398$\pm$0.063\\
MCC& \textbf{0.706$\pm$0.014}& \textbf{0.029$\pm$0.004}& \textbf{0.437$\pm$0.067}\\
\hline
\textit{Emotions}&\multicolumn{3}{c}{\textbf{}} \\
\hline
BR& 0.536$\pm$0.036& 0.240$\pm$0.021& 0.175$\pm$0.036\\
CC& 0.541$\pm$0.034& 0.240$\pm$0.022& 0.184$\pm$0.037\\
ECC& 0.647$\pm$0.038& 0.202$\pm$0.023& 0.283$\pm$0.042\\
MLKNN& 0.529$\pm$0.030& 0.278$\pm$0.022& 0.212$\pm$0.038\\
DMP& 0.607$\pm$0.072& 0.206$\pm$0.023& 0.253$\pm$0.061\\
MCC& \textbf{0.659$\pm$0.081}& \textbf{0.190$\pm$0.023}& \textbf{0.292$\pm$0.084}\\
\hline
\textit{Scene}&\multicolumn{3}{c}{\textbf{}} \\
\hline
BR& 0.672$\pm$0.140& 0.098$\pm$0.037& 0.552$\pm$0.160\\
CC& 0.678$\pm$0.119& 0.109$\pm$0.041& 0.624$\pm$0.117\\
ECC& 0.705$\pm$0.015& 0.094$\pm$0.005& 0.596$\pm$0.016\\
MLKNN& 0.660$\pm$0.114& 0.113$\pm$0.033& 0.553$\pm$0.112\\
DMP& 0.815$\pm$0.077& 0.061$\pm$0.020& 0.704$\pm$0.118\\
MCC& \textbf{0.842$\pm$0.040}& \textbf{0.057$\pm$0.015}& \textbf{0.738$\pm$0.082}\\
\hline
\end{tabular}
\label{tab1}
\end{center}
\end{table}
It is obvious that MCC outperforms the other five algorithms on all metrics. 
For another, as shown in table \ref{cost}, MCC uses less average modality extraction cost than DMP, while other four multi-label algorithms use all the modalities.
\begin{table}[!htbp]
\centering
\caption{Comparison results of the average modality extraction cost, the smaller the better. OTHERS denotes BR, CC, ECC and MLKNN algorithms. The best performance on each dataset is bolded.} 
% Number in parenthesis is the total modality extraction cost. For example, \textit{Herbs}(5.1) means the total modality extraction cost is 5.}
\label{cost}
\begin{tabular}{p{1.3cm}<{\centering} p{2cm}<{\centering}p{2cm}<{\centering} p{2cm}<{\centering}}
\hline
% \diagbox[width=5em,trim=l]{Algorithm}{Datasets} & \textit{Herbs} & \textit{Emotions} & \textit{Scene}\\
\textbf{Algorithm}&\textit{Herbs}& \textit{Emotions}& \textit{Scene}\\
\hline
OTHERS& 5.0$\pm$0.0& 3.0$\pm$0.0& 6.0$\pm$0.0\\
DMP & 3.031$\pm$0.258&1.738$\pm$0.279 &3.202$\pm$0.432\\
MCC & \textbf{2.520$\pm$0.182} & \textbf{1.519$\pm$0.313} & \textbf{2.109$\pm$0.324}\\
\hline
\end{tabular}
\end{table}
% The experimental results show that ECC is the best in the five algorithms, which is in line with our cognitive. We have a certain degree of improvement compared with ECC, while reducing the average extraction cost. 

\section{Conclusion}
Complex objects, i.e., the articles, the images, etc can always be represented with multi-modal and multi-label information.
However, the quality of modalities extracted from various channels are inconsistent. 
Using data from all modalities is not a wise decision.
In this paper, we propose a novel Multi-modal Classifier Chains (MCC) algorithm to improve supplements categorization prediction for MMML problem. 
Experiments in one real-world dataset and two public datasets validate the effectiveness of our algorithm. MCC makes great use of modalities, which can make a convince prediction with many instead of all modalities.
Consequently, MCC reduces modality extraction cost, but it has the limitation of time-consuming compared with other algorithms.
In the future work, how to improve extraction parallelism is a very interesting work.

\section{Acknowledgement}
This paper is supported by the National Key Research and Development Program of China (Grant No. 2016YFB1001102), the National Natural Science Foundation of China (Grant No. 61876080), the Collaborative Innovation Center of Novel Software Technology and Industrialization at Nanjing University.

% References should be produced using the bibtex program from suitable
% BiBTeX files (here: strings, refs, manuals). The IEEEbib.bst bibliography
% style file from IEEE produces unsorted bibliography list.
% -------------------------------------------------------------------------
\bibliographystyle{IEEEbib}
\bibliography{icme2019template}

\begin{thebibliography}{10}

\bibitem{ye2016college}
Han-Jia Ye, De-Chuan Zhan, Xiaolin Li, Zhen-Chuan Huang, and Yuan Jiang,
\newblock ``College student scholarships and subsidies granting: A multi-modal
  multi-label approach,''
\newblock in {\em Data Mining (ICDM), 2016 IEEE 16th International Conference
  on}. IEEE, 2016, pp. 559--568.

\bibitem{lstm1}
Alex Graves and J{\"u}rgen Schmidhuber,
\newblock ``Framewise phoneme classification with bidirectional lstm and other
  neural network architectures,''
\newblock {\em Neural Networks}, vol. 18, no. 5-6, pp. 602--610, 2005.

\bibitem{lstm2}
R~Bertolami, H~Bunke, S~Fernandez, A~Graves, M~Liwicki, and J~Schmidhuber,
\newblock ``A novel connectionist system for improved unconstrained handwriting
  recognition,''
\newblock {\em IEEE Transactions on Pattern Analysis and Machine Intelligence},
  vol. 31, no. 5, 2009.

\bibitem{multilabel}
Min-Ling Zhang and Zhi-Hua Zhou,
\newblock ``A review on multi-label learning algorithms,''
\newblock {\em IEEE transactions on knowledge and data engineering}, vol. 26,
  no. 8, pp. 1819--1837, 2014.

\bibitem{featureselection}
Yvan Saeys, I{\~n}aki Inza, and Pedro Larra{\~n}aga,
\newblock ``A review of feature selection techniques in bioinformatics,''
\newblock {\em bioinformatics}, vol. 23, no. 19, pp. 2507--2517, 2007.

\bibitem{br}
Matthew~R Boutell, Jiebo Luo, Xipeng Shen, and Christopher~M Brown,
\newblock ``Learning multi-label scene classification,''
\newblock {\em Pattern recognition}, vol. 37, no. 9, pp. 1757--1771, 2004.

\bibitem{cc}
Jesse Read, Bernhard Pfahringer, Geoff Holmes, and Eibe Frank,
\newblock ``Classifier chains for multi-label classification,''
\newblock {\em Machine learning}, vol. 85, no. 3, pp. 333, 2011.

\bibitem{etcc}
Yue Peng, Ming Fang, Chongjun Wang, and Junyuan Xie,
\newblock ``Entropy chain multi-label classifiers for traditional medicine
  diagnosing parkinson's disease,''
\newblock in {\em Bioinformatics and Biomedicine (BIBM), 2015 IEEE
  International Conference on}. IEEE, 2015, pp. 856--862.

\bibitem{ldaml}
Yue Peng, Chi Tang, Gang Chen, Junyuan Xie, and Chongjun Wang,
\newblock ``Multi-label learning by exploiting label correlations for tcm
  diagnosing parkinson's disease,''
\newblock in {\em Bioinformatics and Biomedicine (BIBM), 2017 IEEE
  International Conference on}. IEEE, 2017, pp. 590--594.

\bibitem{dimensionreduction}
Sam~T Roweis and Lawrence~K Saul,
\newblock ``Nonlinear dimensionality reduction by locally linear embedding,''
\newblock {\em science}, vol. 290, no. 5500, pp. 2323--2326, 2000.

\bibitem{songandlu}
Xiaonan Song and Haiping Lu,
\newblock ``Multilinear regression for embedded feature selection with
  application to fmri analysis.,''
\newblock in {\em AAAI}, 2017, pp. 2562--2568.

\bibitem{mifs}
Ling Jian, Jundong Li, Kai Shu, and Huan Liu,
\newblock ``Multi-label informed feature selection.,''
\newblock in {\em IJCAI}, 2016, pp. 1627--1633.

\bibitem{lp}
Joseph Wang, Kirill Trapeznikov, and Venkatesh Saligrama,
\newblock ``An lp for sequential learning under budgets,''
\newblock in {\em Artificial Intelligence and Statistics}, 2014, pp. 987--995.

\bibitem{dag}
Joseph Wang, Kirill Trapeznikov, and Venkatesh Saligrama,
\newblock ``Efficient learning by directed acyclic graph for resource
  constrained prediction,''
\newblock in {\em Advances in Neural Information Processing Systems}, 2015, pp.
  2152--2160.

\bibitem{dmp}
Yang Yang, De-Chuan Zhan, Ying Fan, and Yuan Jiang,
\newblock ``Instance specific discriminative modal pursuit: A serialized
  approach,''
\newblock in {\em Asian Conference on Machine Learning}, 2017, pp. 65--80.

\bibitem{end2end}
Wei Li, Zheng Yang, and Xu~Sun,
\newblock ``Exploration on generating traditional chinese medicine prescription
  from symptoms with an end-to-end method,''
\newblock {\em arXiv preprint arXiv:1801.09030}, 2018.

\bibitem{gini}
Leo Breiman,
\newblock {\em Classification and regression trees},
\newblock Routledge, 2017.

\bibitem{begum2015data}
Shemim Begum, Debasis Chakraborty, and Ram Sarkar,
\newblock ``Data classification using feature selection and knn machine
  learning approach,''
\newblock in {\em Computational Intelligence and Communication Networks (CICN),
  2015 International Conference on}. IEEE, 2015, pp. 811--814.

\bibitem{adadelta}
Matthew~D Zeiler,
\newblock ``Adadelta: an adaptive learning rate method,''
\newblock {\em arXiv preprint arXiv:1212.5701}, 2012.

\bibitem{emotions}
Konstantinos Trohidis, Grigorios Tsoumakas, George Kalliris, and Ioannis~P
  Vlahavas,
\newblock ``Multi-label classification of music into emotions.,''
\newblock in {\em ISMIR}, 2008, vol.~8, pp. 325--330.

\bibitem{mlknn}
Min-Ling Zhang and Zhi-Hua Zhou,
\newblock ``Ml-knn: A lazy learning approach to multi-label learning,''
\newblock {\em Pattern recognition}, vol. 40, no. 7, pp. 2038--2048, 2007.

\bibitem{fmeasure2}
Krzysztof Dembczynski, Arkadiusz Jachnik, Wojciech Kotlowski, Willem Waegeman,
  and Eyke H{\"u}llermeier,
\newblock ``Optimizing the f-measure in multi-label classification: Plug-in
  rule approach versus structured loss minimization,''
\newblock in {\em International Conference on Machine Learning}, 2013, pp.
  1130--1138.

\end{thebibliography}

\end{document}